\documentclass[letterpaper]{article} 
\usepackage{aaai25}  
\usepackage{times}  
\usepackage{helvet}  
\usepackage{courier}  
\usepackage[hyphens]{url}  
\usepackage{graphicx} 
\usepackage{natbib}  
\usepackage{caption} 

\usepackage{algorithm}
\usepackage{algorithmic}
\usepackage{mathrsfs}
\usepackage{amsmath}
\usepackage{amssymb}
\usepackage{algorithm}
\usepackage{algorithmic}

\usepackage{newfloat}
\usepackage{listings}
\DeclareCaptionStyle{ruled}{labelfont=normalfont,labelsep=colon,strut=off} 
\floatstyle{ruled}
\newfloat{listing}{tb}{lst}{}
\floatname{listing}{Listing}
\title{StyO: Stylize Your Face in Only One-shot}
\author {
    Bonan Li\textsuperscript{\rm 1}\thanks {Equal contribution},
    Zicheng Zhang\textsuperscript{\rm 1}\footnotemark[1],
    Xuecheng Nie\textsuperscript{\rm 2},
    Congying Han\textsuperscript{\rm 1}\thanks{Corresponding author},\\
    Yinhan Hu\textsuperscript{\rm 1},
    Xinmin Qiu\textsuperscript{\rm 1},
    Tiande Guo\textsuperscript{\rm 1}
}
\affiliations {
    \textsuperscript{\rm 1} University of Chinese Academy of Sciences\quad
    \textsuperscript{\rm 2} MT Lab, Meitu Inc.\\

    \{libonan, hancy, tdguo\}@ucas.ac.cn,\ 
    \{zhangzicheng19, huyinhan16, qiuxinmin21\}@mails.ucas.ac.cn,\ nxc@meitu.com
    
}

\usepackage{bibentry}

\begin{document}

\maketitle

\begin{abstract}
This paper focuses on face stylization with a single artistic target. Existing works for this task often fail to retain the source content while achieving geometry variation. Here, we present a novel StyO model, \textit{i.e.} \textbf{Sty}lize the face in only \textbf{O}ne-shot, to solve the above problem. In particular, StyO exploits a disentanglement and recombination strategy. It first disentangles the content and style of source and target images into identifiers, which are then recombined in a cross manner to derive the stylized face image. In this way, StyO decomposes complex images into independent and specific attributes, and simplifies one-shot face stylization as the combination of different attributes from input images, thus producing results better matching face geometry of target image and content of source one. StyO is implemented with latent diffusion models (LDM) and composed of two key modules: 1) Identifier Disentanglement Learner (IDL) for disentanglement phase. It represents identifiers as contrastive text prompts, \textit{i.e.} positive and negative descriptions. And it introduces a novel triple reconstruction loss to fine-tune the pre-trained LDM for encoding style and content into corresponding identifiers; 2) Fine-graind Content Controller (FCC) for recombination phase. It recombines disentangled identifiers from IDL to form an augmented text prompt for generating stylized faces. In addition, FCC also constrains the cross-attention maps of latent and text features to preserve source face details in results. The extensive evaluation shows that StyO produces high-quality images on numerous paintings of various styles and outperforms the current state-of-the-art. 
\end{abstract}

%
\vspace{-4mm}
\section{Introduction}
\label{sec:intro}
Face stylization aims at automatically creating personalized artistic portraits from face photographs. It is a fundamental problem in vision and graphics communities~\cite{ErgunAkleman1997MakingCW,Kim2019UGATITUG,Han2021ExemplarBased3P,Song2021AgileGANSP} and widely applied in creative industries such as social media avatars, films, advertising, etc. Recently, one-shot solution for this task has drawn lots of attention, due to its significant feature of data-efficiency, which makes the deployment easier in practice. 

Existing works \cite{Chong2021JoJoGANOS,Zhu2021MindTG,Zhang2022GeneralizedOD,Zhang2022TowardsDA} consider one-shot face stylization as a style transfer problem~\cite{LeonAGatys2016ImageST}, \textit{e.g.}, transferring face style of a single target image into face of the source image, and solve this problem with Generative Adversarial Networks (GANs)~\cite{Goodfellow2014GenerativeAN,karras2019style}. Despite of their success, we observe that even state-of-the-art methods still fail to achieve reasonable style (\textit{e.g.}, geometry variations) while maintain source content (\textit{e.g.}, facial and hair colors), as shown in Figure~\ref{fig:teaser}. Note that these features are delineated as fundamental attributes warranting consideration \cite{KaidiCao2018CariGANsUP,Gong2020AutoToonAG,YichunShi2018WarpGANAC,YifangMen2022DCTNetDT,Han2021ExemplarBased3P,Song2021AgileGANSP,Liu2021BlendGANIG}. This is caused by two main reasons. First, these methods heavily relied on pre-trained GAN models on large real face dataset \cite{karras2019style}, thus hard to match styles of faces, especially geometry changes, from another domain with one-shot guidance, due to capability limitation of GANs for producing distribution variations. Second, they entangle the style and content information of facial images together in the latent space, making current encoding and  inversion techniques difficult to derive accurate latent code representing the stylized faces. It is important to address above issues for facilitating usage of relevant techniques in real world.

Motivated by this, we propose a novel StyO model in this paper for pushing forward the frontier of the task of \textbf{Sty}lizing the face in only \textbf{O}ne-shot. For improving the generation variety, StyO exploits Diffusion Denosing Probabilistic Models (DDPMs) \cite{DDPM,ddim,ldm} instead of GANs, given the superiority of DDPM to produce high-quality images in wild range of distribution.  For deriving accurate representation of stylized faces, StyO leverages a disentanglement and recombination strategy. Specifically, StyO factorizes the style and content information of source and target images into different identifiers, then fuse them in a cross manner to form prompts, which plays as representations of stylized faces to drive DDPMs to generate reasonable images. In this way, StyO decomposes complex images into independent and specific attributes, and simplifies one-shot face stylization as combination of different attributes from input images, thus producing results better matching face style of target image and content of source one. 

\begin{figure}
\begin{center}
\includegraphics[width=1\linewidth]{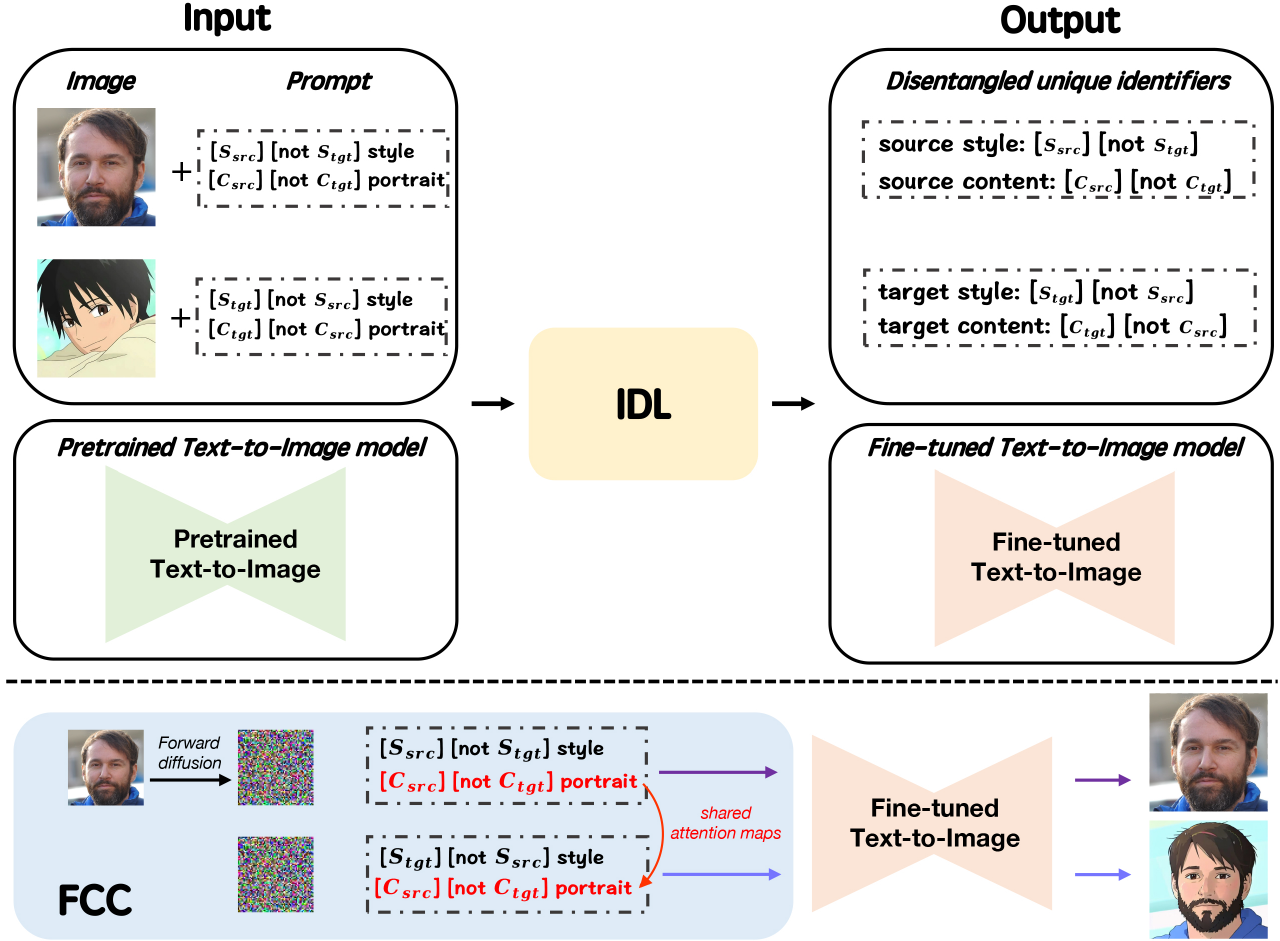}
\end{center}
\vspace{-4mm}
   \caption{Overview of the proposed StyO for one-shot face stylization. StyO consists of two core modules: Identifier Disentanglement Learner (IDL) and Fine-grained Content Controller (FCC). IDL learns to disentangle style and content information of source and target images into different identifiers. Then, FCC recombine disentangled identifiers for fusing source content and target style to generate the stylized facial images.  
   }
\label{fig:highlevel}
\vspace{-6mm}
\end{figure}

In particular, StyO is implemented with the efficient Latent Diffusion Models (LDMs) \cite{ldm}. It is composed of two core modules: 1) Identifier Disentanglement Learner (IDL). IDL aims to disentangle the style and content of an image into different identifiers. Here, IDL represents identifiers as text descriptors for fully leveraging the powerful capability of text-based image generation of LDMs. Then, IDL defines contrastive disentangled prompt template as text descriptions for input images with content and style identifiers, \textit{e.g.}, ``a drawing with [\textit{Source/Target Style Identifier}][not \textit{Target/Source Style Identifier}] style of [\textit{Source/Target Content Identifier}][not \textit{Target/Source Content Identifier}] of portrait'' for source and target images, respectively. In addition, we introduce an auxiliary prompt template with only source and target style identifiers to describe an auxiliary image set with same style of source one. This helps style and content identifiers correctly to represent corresponding attributes of images, and also avoid the risk of relating style/content information with other words in prompts rather than corresponding identifiers. Given prompts defined above, StyO builds text-image pairs to fine-tune the pre-trained LDMs for injecting attributes of images into identifiers, thus achieving the disentanglement goal. 2) Fine-grained Content Controller (FCC). FCC aims to recombine style and content identifiers from IDL to generate stylized face images. Specifically, FCC constructs prompt template as ``a drawing with [\textit{Target Style Identifier}][not \textit{Source Style Identifier}] style of [\textit{Source Content Identifier}][not \textit{Target Content Identifier}] of portrait'' to describe stylized faces with style from target and content from source, which is used as condition to LDMs for deriving stylization results. However, we find that only with the above reconstructed prompt will cause the loss of fine-grained details of source image, such as head pose, hair color, beard style, etc, due to the randomness feature of diffusion models yielding less control on diversity. To solve this issue, FCC presents a new manipulation mechanism on attention maps to improve  controllability. This is inspired by the fact that attention maps from cross attention layers are semantically correlated with texts in prompts. Hence, FCC extracts attention maps for source content and uses them to replace that for stylized one, which effectively controls fine-grained details of results. In addition, FCC proposes to augment prompts by repeating identifiers. This simple augmentation strategy further successfully advances generation quality.
By this design as shown in Figure~\ref{fig:highlevel}, StyO can produce faces with suitable target style while maintaining source content.

Extensive experiments show that the proposed StyO model can produce surprisingly good quality of facial images on various styles in the one-shot manner. It also outperforms state-of-the-art models quantitatively. Our contributions are summarized in four folds: First, to the best knowledge, we are the first to deploy diffusion models for tackling one-shot face stylization; Second, we present a novel StyO model with disentanglement and recombination strategy, which effectively solves core issues of prior works; Third, we present a classical way to encoding different attributes of images into multiple identifiers; Fourth, our StyO model sets new state-of-the-art for one-shot face stylization.
\vspace{-4mm}
\section{Related Work}
\label{related}
\noindent\textbf{{\textit{Diffusion model}}} recently breaks the long-time dominance of generative adversarial networks (GANs) \cite{Goodfellow2014GenerativeAN} on generative tasks. Since  \citet{JaschaSohlDickstein2015DeepUL}  first propose the denoising diffusion probabilistic models (DDPM) \cite{DDPM,AlexNichol2021ImprovedDD,PrafullaDhariwal2021DiffusionMB} to form the image generation as a Markov Process, the community is inspired  to investigate the generative model from non-Markov process \cite{ddim}, score matching \cite{YangSong2020ImprovedTF}, and stochastic differential equation \cite{Song2020ScoreBasedGM}. Besides, many sampling techniques are proposed to speedup \cite{ddim,LupingLiu2023PseudoNM,TeroKarras2022ElucidatingTD} or condition \cite{PrafullaDhariwal2021DiffusionMB,JonathanHo2023ClassifierFreeDG} the inference. Recent text-to-image diffusion models \cite{AlexNichol2023GLIDETP,AdityaRamesh2023HierarchicalTI,ldm,imagen} trained on
extremely large-scale data have exhibited remarkable ability to generate diverse, creative images controlled by text prompts.  \cite{RinonGal2023AnII,dreambooth,nulltext,imagic,Hertz2022PrompttoPromptIE,EI} shows that these models can edit images provided by users with a text prompt effectively, thereby enabling text-based stylization.
However, compared with the fine-grained details provided by an artistic exemplar,  the text prompt only provides a rough description of desirable style, thereby it is hard to obtain acceptable results with previous methods. To our best knowledge, StyO is the first to explore the exemplar-guided fine-style transfer in context of text-to-image diffusion model. 

\noindent\textbf{{\textit{Face stylization}}} has been widely studied in the field of Neural Style Transfer (NST), image-to-image translation (I2IT), and GAN adaptation. 
NST methods match the feature distribution \cite{LeonAGatys2016ImageST,YanghaoLi2017DemystifyingNS,LiPan2019OptimalTO,Kolkin2019StyleTB,Kalischek2021InTL} to transfer artistic texture. These methods can achieve pleasing results when the artistic portraits have an intense color or texture pattern, but usually underperform in styles involving significant geometric distortion of facial features, such as anime and cartoons. I2IT methods \cite{PhillipIsola2016ImagetoImageTW,JunYanZhu2017UnpairedIT,Liu2017UnsupervisedIT,wang2018pix2pixHD,JunhoKim2019UGATITUG} learn a mapping from a large-scale dataset to translate an input image from a source domain to a target domain. 
\citet{YifangMen2022DCTNetDT} design specific modules and losses based on facial knowledge (\textit{e.g.}, landmarks, expression) to facilitate the face stylization. These methods can achieve reasonable texture and geometric variation when the training dataset includes more than hundreds of artistic portraits. However, collecting such extensive images for a specific style is not non-trivial, thus limiting the usage of I2IT in practice. Recently, the works about domain adaptation of GANs \cite{Ojha2021FewshotIG} provide a creative way to solve the task. Since the pre-trained StyleGAN \cite{karras2019style} forms a generative prior of natural faces, an artistic portrait, which belongs to the proximal domain of natural face, can guide the adaptation of StyleGAN \cite{Zhu2021MindTG,Gal2021StyleGANNADACD,Chong2021JoJoGANOS,Zhang2022GeneralizedOD} to generate artistic portraits. Thereafter, given a face photograph, a latent code can be found by inversion techniques \cite{Xia2021GANIA} and fed into adapted StyleGAN to generate the stylized image. However, the strong facial prior of pre-trained StyleGAN limits the geometric variation, and compressing face images into latent code also drops the content partially. Hence these methods perform not well when implementing artistic images of large geometric deformations and face photographs out of the facial prior. In contrast, StyO properly utilizes the potent generative and reconstructive abilities of LDM to alleviate these problems.

\begin{figure*}
\begin{center}
\includegraphics[width=0.9\linewidth]{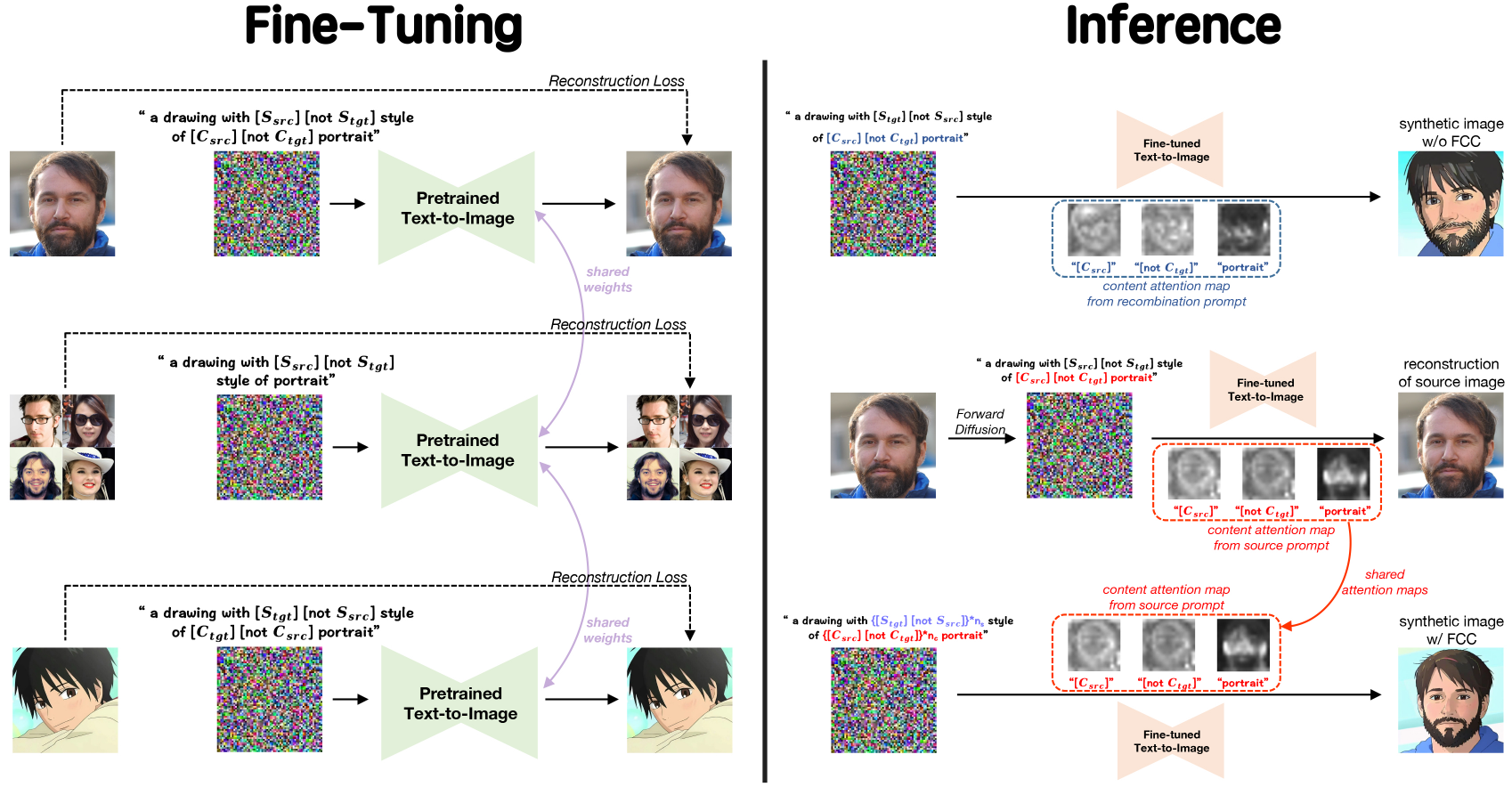}
\end{center}
\vspace{-4mm}
   \caption{Pipeline of the proposed StyO. Left panel: The training phase. StyO constructs text-image pairs for source, target images and auxiliary image set with a contrastive disentangled prompt template. Then, StyO fine-tunes a pre-trained latent diffusion model in the \textit{one-shot} manner.  In this way, StyO injects attributes of images into different identifiers. Right panel: The inference phase. Given style and content identifiers, StyO recombines them to form text prompt to generating stylized face. To maintain fine-grained details, StyO extracts attention maps for source content and uses them to replace that for stylized one. This process yields faces with suitable target style while maintains good source content.
   }
\label{fig:framework}
\vspace{-7mm}
\end{figure*}
\vspace{-4mm}
\section{Methodology}
\vspace{-2mm}
\subsection{Preliminaries}
\vspace{-1mm}
Text-to-image diffusion models~\cite{ldm,imagen} are a family of generative models to map a noise latent $\epsilon$ and a text prompt $\mathcal{P}$ to an image $x_0$. 
In this paper, we focus on the pre-trained latent diffusion model (LDM)~\cite{ldm}, which is widely used as a base model in recent research.
LDM first encodes the image $x^{0}$ into latent space with a pre-trained encoder $\mathcal{E}$, \textit{i.e.}, $z^0 = \mathcal{E}(x^{0})$, while $x^{0}$ can be restored by means of a pre-trained decoder $\mathcal{D}$. It then performs the destructive and reverse processes in the latent space.
Specifically, the destructive process is known as a fixed-length Markov chain starting from $z^{0}$ to $z^{T}$ with a Gaussian transition at each step. For each $t \in \{0,...,T\}$, the transition holds that
\begin{equation}
\label{eq:destruct}
z^{t} = \sqrt{\alpha^{t}}z^{0}+\sqrt{1-\alpha^{t}}\epsilon,\  \epsilon\sim N(0,I),
\end{equation}
where $\{\alpha^{t}\}_{t=0}^{T}$ is a monotonic, strictly decreasing sequence, in which $\alpha^{0}=1$ and $\alpha^{T}$ vanishes as $T$ increases. This ensures $z^T$ roughly follows an isotropic Gaussian distribution. In contrast, the reverse process involves sequential Gaussian sampling given by
\begin{equation}
    z^{t-1} =\mu(z^{t},\epsilon_{\theta},t)+\sigma(t) w,\   w\sim N(0,I).
\end{equation}
Here, $\mu$ and $\sigma$ are pre-defined functions to compute parameters of Gaussian distribution. $\mu$ is parameterized by $\epsilon_{\theta}$ that is a time-conditioned UNet~\cite{unet}  equipped with attention mechanism and trained to achieve the objective 
\begin{equation}
\label{eq:train}
\mathop{\mathrm{min}}_{\theta}{\mathbb{E}}_{z^{0}, \epsilon,t}||\epsilon-\epsilon_{\theta}(z^{t},t,\gamma)||^2_2.
\end{equation}
$\gamma$ is the text embedding of $\mathcal{P}$ mapped by a frozen model $\phi$, \textit{e.g.}, the CLIP text encoder~\cite{clip}. The reverse process ensures the final sample $z^{0}$ follows the distribution of latents, and the generated image aligns with the prompt.

\vspace{-3mm}
\subsection{Framework of StyO}
\vspace{-1mm}
Given a source image $x_{src}$ in natural face domain, and a target image $x_{tgt}$ in artistic portrait domain, our objective is to synthesize an image with the source content as well as the target style. Building upon the state-of-the-art LDM \cite{ldm}, we propose a novel approach named StyO, which consists identifier disentanglement learner (IDL) and fine-grained content controller (FCC) to tackle the task. In general, StyO learns disentangled identifiers representing content and style via IDL, and enables the retention of fine-grained content details through FCC when inferring the stylized image with the prompt comprising recombined identifiers. A detailed sketch of our proposed StyO is illustrated in Figure~\ref{fig:framework}.

\vspace{-2mm}
\subsubsection{\textbf{Identifier Disentanglement Learner}} 
In order to extract the content and style information from given images, we design a novel contrastive disentangled prompt template with positive-negative identifiers and information descriptors to construct the text label for each image. These image-text pairs will be used to fine-tune LDM with a triple reconstruction loss, such that the content and style are encoded into identifiers.

\noindent\textbf{{\textit{Contrastive Disentangled Prompt Template.}}} As shown in Figure~\ref{fig:framework}, we first label each image with a novel contrastive disentangled prompt template: ``a drawing with [$S^{+}$][$S^{-}$] style of [$C^{+}$][$C^{-}$] portrait'', where ``style'' and ``portrait'' are the information descriptor, ``[$S^{+}$]''/``[$S^{-}$]'' is a positive/negative unique identifier linked to the style, and ``[$C^{+}$]''/``[$C^{-}$]'' is a positive/negative unique identifier linked to the content of subject. Specifically, the positive identifiers are expected to hold semantic meaning corresponding to the image. With regard to the source/target image, we use ``[$S_{src}$]''/``[$S_{tgt}$]'' and ``[$C_{src}$]''/``[$C_{tgt}$]''  for ``[$S^{+}$]'' and ``[$C^{+}$]'', respectively. In contrast,
the negative identifier is the nomenclature proposed to represent elements that are not presented in the image. We cooperate the negator, \textit{e.g.}, ``not'', ``without'' or ``except'', with above positive identifiers to construct negative identifiers. For instance, we set ``[$S^{-}$]'' as ``[$not\ S_{tgt}$]'', and ``[$C^{-}$] as ``[$not\ C_{tgt}$]'' for the negative identifiers in the prompt of source image. Such design of contrastive prompt highlights the significant visual disparity between the source and target images, and will facilitate the disentangled learning by the effective exploitation of the limited given data.

In addition to the task-relevant data $x_{src}$ and $x_{tgt}$, we introduce a \textit{no-cost} auxiliary image set $x_{aux}$ to further enhance the disentanglement, in which about 200 natural faces are randomly sampled from FFHQ \cite{karras2019style} and labeled with a same prompt ``a drawing with [$S_{src}$][$not\ S_{tgt}$] style of portrait''. The contrast between prompts of $x_{aux}$ and $x_{src}$ helps to encode source content into identifiers ``[$C^{+}$]''/``[$C^{-}$]'' instead of the descriptor ``portrait'', and strengthen the style distinction
between $x_{src}$ and $x_{tgt}$. To sum up, we list prompts corresponding to different images or set as:

- Prompt of source image ($\mathcal{P}_{src}$) : ``a drawing with [$S_{src}$][$not\ S_{tgt}$] style of [$C_{src}$][$not\ C_{tgt}$] portrait''.

- Prompt of target image ($\mathcal{P}_{tgt}$): ``a drawing with [$S_{tgt}$][$not\ S_{src}$] style of [$C_{tgt}$][$not\ C_{src}$] portrait''.

- Prompt of auxiliary image set ($\mathcal{P}_{aux}$): ``a drawing with \newline [$S_{src}$][$not\ S_{tgt}$] style of portrait''.

Note that, different from ~\cite{dreambooth} using rare words in T5-XXL tokenizer, we simply use firearms (\textit{e.g.}, ``sks'', ``aug'' and ``ak47'') as  identifiers and also achieve phenomenal results.
\vspace{-1mm}

\noindent\textbf{{\textit{Triple Reconstruction Loss.}}}
Fine-tuning of diffusion models is a powerful technique relevant to many use cases, \textit{e.g.}, image editing~\cite{unitune,imagen,dreambooth} and image-to-image translation~\cite{pallnedd}. To make the identifiers work as intended, we opt for this simple approach with the text-image pair $\{(\mathcal{P}_{i}, x_{i})\}_{i\in \{src, tgt, aux\}}$. We tune the text-to-image model with a triple reconstruction loss, thereby injecting different information to individual identifiers:
\begin{equation}
\label{eq:train}
\begin{aligned}
&\mathop{\mathrm{min}}_{{\theta}}{\mathbb{E}}_{z_{src},z_{tgt},z_{aux}, \epsilon,t}||\epsilon-\epsilon_{\theta}(z_{src}^{t},t,\gamma_{src})||\\
&+||\epsilon-\epsilon_{\theta}(z_{tgt}^{t},t,\gamma_{tgt})||+||\epsilon-\epsilon_{\theta}(z_{aux}^{t},t,\gamma_{aux})||,
\end{aligned}
\end{equation}
where $z_{src}=\mathcal{E}(x_{src})$, 
$\gamma_{src}=\phi(\mathcal{P}_{src})$, $z_{src}^{t}$ is the noisy latent according to Eq.~\eqref{eq:destruct}, and same goes for other symbols. Note, we simultaneously
fine-tune the text encoder $\phi$ 
to achieve better results. By implementing IDL, we can obtain a tuned model and  disentangled identifiers biased to given images.

\vspace{-2mm}
\subsubsection{\textbf{Fine-grained Content Controller}}

The proposed IDL allows for the effective fusion of source identity and artistic style via the recombination of identifiers:

- Prompt of stylized portrait ($\mathcal{P}_{sty}$): ``a drawing with \newline [$S_{tgt}$][$not\ S_{src}$] style of [$C_{src}$][$not\ C_{tgt}$] portrait'' 

Nonetheless, as depicted in Figure \ref{fig:framework}, the synthesis that is highly correlated with the source face, exhibits noticeable diversity in terms of poses and some facial features from source image. Although this diversity can offer additional customization options, it would be desirable for the model to consistently produce artistic portraits that accurately capture the source face.  To alleviate this problem, we introduce a Fine-grained Content Controller (FCC) module that further transfers delicate details of source image into syntheses.  In technique, FCC devises the cross attention control and augments text prompt to ensure content attention maps of the stylized portrait correspond closely to those of the source face, as depicted in the right panel of Figure \ref{fig:framework}.

\noindent\textbf{{\textit{Cross Attention Control.}}} LDM utilizes a cross attention mechanism to modify the latent features based on the conditional text features. Here we slightly abuse some notations: given the text embedding $\gamma \in \mathbb{R}^{s\times d_{1}}$, latent  features $z \in \mathbb{R}^{hw\times d_{2}}$, where $s$ denotes the number of tokens,    $h$ and $w$ are the height and width of latent feature, and $d_1$ and $d_2$ are the dimensions of features, respectively. The cross attention first computes query $Q = zW^q$, key $K = \gamma W^k$, and
value $V = \gamma W^v$. Then latent features are updated by 
\begin{small}
\begin{equation}
\label{eq:attention_1}
M(z,\gamma)=\mathop{softmax}\left(\frac{QK^{T}}{\sqrt{d}}\right),
\end{equation}
\begin{equation}
\label{eq:attention_2}
\text{Attn}(z,\gamma)= M(z,\gamma)V,
\end{equation}
\end{small}
where $d$ is the dimension of key and query feature.  

\citet{Hertz2022PrompttoPromptIE} find that the attention map $M$ in Eq.~\eqref{eq:attention_1} has a strong influence on the  spatial layout of syntheses. In the face domain, we are inspired to control the pose and facial details of stylized portraits by constraining the attention mask of syntheses to that of source image. Specifically, considering the feed-forward process of LDM prompted by $\mathcal{P}_{src}$ and $\mathcal{P}_{sty}$, their latent features and text features are ($z_{sty}$, $z_{src}$) and  ($\gamma_{sty}$,$\gamma_{src}$), respectively. We modify the cross attention to a condition form:
\begin{equation}
\label{eq:swap mask}
\begin{aligned}
&\text{Attn}(z_{sty}, \gamma_{sty}; z_{src},\gamma_{src})= M_{ctr}V_{sty},\\
&\text{where} \  M_{ctr}^{i}= M_{src}^{i} \ \text{if}\  i \in \text{Content Index} \  \text{else} \  M_{sty}^{i}.
\end{aligned}
\end{equation}
$M_{sty}$ and $M_{src}$ are the attention maps computed by the corresponding text feature and latent feature, and the superscript $i$ denotes the $i$-th column of matrix, and the $\text{Content Index}$ includes the indexes of ``[$C_{src}$]'',  ``[$not\ C_{tgt}$]'' and  ``portrait'' in $\mathcal{P}_{src}$. Note that the tokens of $\mathcal{P}_{src}$ and $\mathcal{P}_{sty}$ have a same length, thus FCC indeed swaps the attention maps of those content identifiers to pull the content of the stylized portrait towards source image.

\noindent\textbf{{\textit{Augmented Text Prompt.}}} The text prompt is a critical factor in controlling the content of syntheses for LDM, and improving its quality usually enhances the semantic details of syntheses directly. Leveraging the disentangled identifiers obtained from IDL, we introduce an augmented text prompt that offers greater customization and control over stylized portraits. Given a identifier ``[C]'' and $n \in \mathbb{Z}^{+}$, we define the augmented identifier ``[C]''$*n$ as repeating the identifier $n$ times. Augmented text prompts $\mathcal{P}^{aug}_{src}$ and $\mathcal{P}^{aug}_{sty}$ comprises augmented identifiers:

- Augmented prompt of source face ($\mathcal{P}^{aug}_{src}$): ``a drawing with ([$S_{src}$][$not\ S_{tgt}$])*$n_s$ style of ([$C_{src}$][$not\ C_{tgt}$])*$n_c$ portrait''.

- Augmented prompt of stylized portrait ($\mathcal{P}^{aug}_{sty}$): ``a drawing with ([$S_{tgt}$][$not\ S_{src}$])*$n_s$ style of ([$C_{src}$][$not\ C_{tgt}$])*$n_c$ portrait''.

Instead of $\mathcal{P}_{src}$ and $\mathcal{P}_{sty}$, we use prompt $\mathcal{P}^{aug}_{src}$ to reconstruct the source face and $\mathcal{P}^{aug}_{sty}$ to generate the stylized portrait with the aid of cross attention control. It is important to note that two hyper-parameters, $n_s$ and $n_c$, are introduced to control the degree of stylization and identity preservation. Compared with the single identifier as used during training, a balance of $n_s$ and $n_c$ will result in better effects. Actually, it is not complicated to determine the optimal values for the two hyper-parameters. Through experiments, we find setting $n_s$ and $n_c$ to small values ($\leq 3$) is adequate to produce satisfactory results.

\noindent\textbf{{\textit{Inference with FCC.}}}
On this basis, the process of generation with FCC can be summarized as follows:  We carry out destructive process Eq. \eqref{eq:destruct} to obtain the noisy latent $z_{src}^{T}$ of source image. Note that compared with random Gaussian noise, $z_{src}^{T}$ provides a better prior to restore $x_{src}$. Then we perform the reverse diffusion starting from $z_{src}^{T}$ and prompted by $\mathcal{P}^{aug}_{src}$. Meanwhile, their attention maps $M_{sty}$ in all cross attention layers are recorded as Eq. \eqref{eq:attention_1}. Once again, we perform the reverse diffusion starting from $z_{src}^{T}$ and prompted by $\mathcal{P}^{aug}_{sty}$ to synthesize the stylized portrait, while it implements the conditional cross attention as Eq.~\eqref{eq:swap mask} to fetch fine-grained details from source image. 
\begin{figure*}
\begin{center}
\includegraphics[width=1\linewidth]{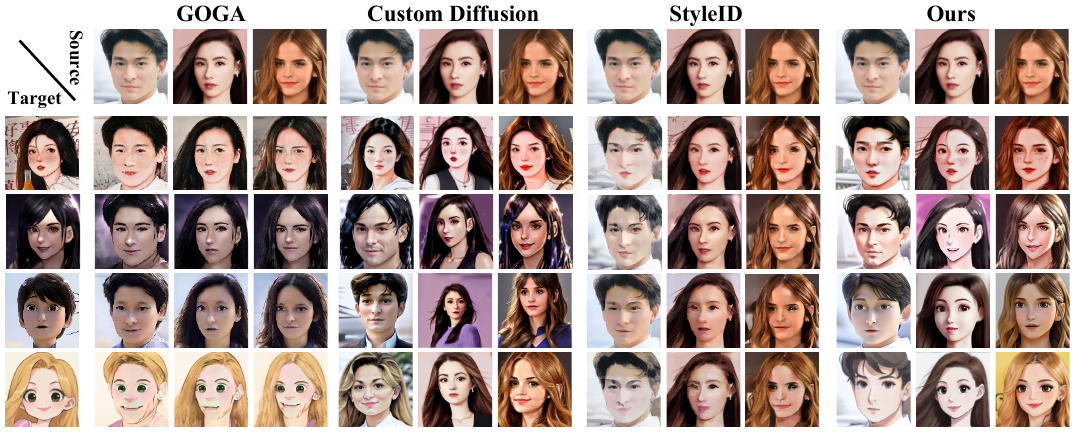}
\end{center}
\vspace{-4mm}
   \caption{Qualitative comparison of our StyO with GOGA~\cite{Zhang2022GeneralizedOD}, Custom Diffusion~\cite{kumari2023multi}, and StyleID~\cite{zstar}. Better see in color and 2x zoom.}
\label{fig:sota}
\vspace{-6mm}
\end{figure*}
\vspace{-3mm}

\begin{figure}
\begin{center}
\includegraphics[width=\linewidth]{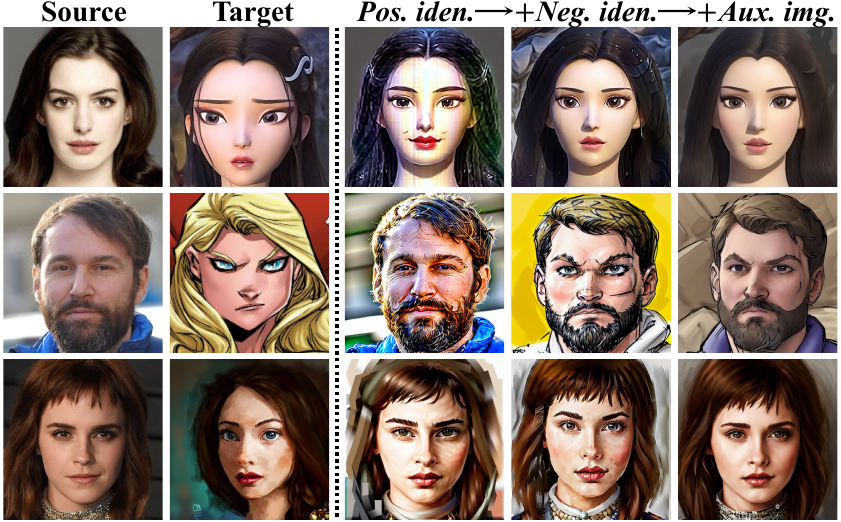}
\end{center}
   \vspace{-4mm}
   \caption{Ablation of Contrastive Disentangled Prompt Template. The arrows indicate the changes made to the current prompt template by sequentially plusing positive identifiers, negative identifiers, and auxiliary image set. Better see in color and 2x zoom.}
\label{fig:aba1}
\vspace{-7mm}
\end{figure}
\section{Experiments}
\label{sec: exp}
\begin{figure*}
\begin{center}
\includegraphics[width=1\linewidth]{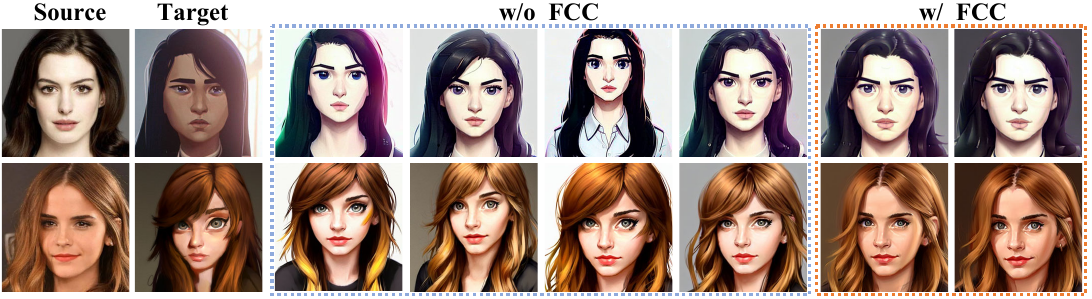}
\end{center}
\vspace{-4mm}
   \caption{Ablation of Fine-grained Content Controller. Better see in color and 2x zoom.}
\label{fig:aba2}
\vspace{-4mm}
\end{figure*}

\begin{table}
\scriptsize
\begin{center}

\begin{tabular}{c|c|c|c}
\hline
Model    & Identity$\uparrow$  & Geometry$\uparrow$ & Texture$\uparrow$  \\ \hline

GOGA~\cite{Zhang2022GeneralizedOD}     & 0.11    & 0.14     & 0.23     \\
Custom Diffusion~\cite{kumari2023multi}  & 0.09    & 0.27     & 0.11     \\
StyleID~\cite{zstar} & 0.38    & 0.12     & 0.28    \\
\textbf{Ours}     & \textbf{0.42}    & \textbf{0.47}     & \textbf{0.38}     \\
\hline
\end{tabular}
\caption{Quantitative comparison of our StyO and state-of-the-arts based on the user study. Please see details in text.}
\label{tab:user}
\end{center}
\vspace{-8mm}
\end{table}
StyO utilizes IDL to fine-tune LDM, and generates stylized portraits by means of FCC. We first conduct both qualitative and quantitative experiments to compare our method with previous state-of-the-art methods of one-shot face stylization. Additionally, we perform the ablation study to demonstrate the effectiveness of the proposed IDL and FCC.

\begin{figure*}[h]
\begin{center}
\includegraphics[width=1\linewidth]{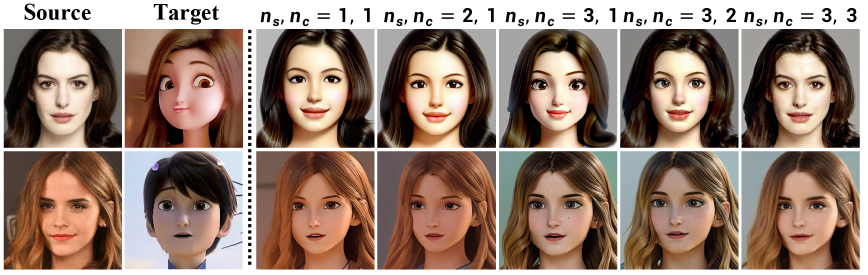}
\end{center}
\vspace{-4mm}
   \caption{Ablation of hyperparameter in augmented text prompt. Better see in color and 2x zoom.}
\label{fig:aba3}
\vspace{-7mm}
\end{figure*}

\vspace{-2mm}
\subsection{Comparison with SOTA methods}
\noindent\textbf{{\textit{Baselines.}}}
We compare StyO with two recent representative state-of-the-arts of  GOGA~\cite{Zhang2022GeneralizedOD}, which tune the pre-trained StyleGAN to the target domain and stylize faces via GAN inversion \cite{Xia2021GANIA}. We also report the results from a concurrent diffusion-based stylization Custom Diffusion~\cite{kumari2023multi} and StyleID~\cite{zstar} to show the advantage of StyO. All these methods are implemented by their official codes. We refrain from comparing with InstantID~\cite{wang2024instantid} due to its lack of capability in extracting style from target images. The artistic portraits are token from AAHQ dataset \cite{blengan}, while the natural faces employ photos of celebrities.

\noindent\textbf{{\textit{Qualitative Comparison.}}}
Figure~\ref{fig:sota} shows the qualitative results. The top row images are the source natural faces and the leftmost images are the target artistic portraits. From the results, we can draw the following conclusions: 1) Our method not only achieves facial identity consistency between the source and output images, but also highly preserves local details such as hair and eye color, which is a significant improvement over other methods. In contrast, other methods often fail to generate images that are faithful to the identity information of source images and transfers unexpected colors in some cases. 2) Our method captures the desired geometric variations in the target image, such as big eyes and round face. Instead of using GAN based generative model, we use potential text-image diffusion model to successfully capture the geometric variation in reference image and inject it to unique identifier. Obviously, other methods can not associate big eyes and round faces as consistent geometric variations and smoothly introduce both of these into the source image. 
(3) Our method produces syntheses that flexibly acquire vivid local stroke characteristics and overall appearance, resulting in high-quality images with sharp and realistic high-frequency details. In contrast, the results produced by other methods tend to be fuzzy. 

\noindent\textbf{{\textit{Quantitative Comparison.}}}
\label{sec:sota}
As portrait stylization is often regarded as a subjective task, we resort to user studies to evaluate which method generates results that are most favored by humans. We conducted three user studies on the results in terms of the identity, geometric and texture. In the first study, participants are asked to select the stylized images best preserve the identity of source image. In the second study, participants are asked to point out the best stylized images with reasonable exaggerated geometric deformation. In the third study, we encourage participants to touch the image with the finest texture and less distorted artifacts. We receive 150 answers on 50 source-target pairs in total for each study. As shown in Table~\ref{tab:user}, over 0.38\% of our results are selected as the best in both three metrics, which proves a significant advantage in stylization.

\vspace{-3mm}
\subsection{Ablation Study}
\label{sec:aba}

\noindent\textbf{{\textit{Effect of Contrastive Disentangled Prompt Template.}}}
Our proposed contrastive disentangled prompt template acts as a powerful tool to disentangle and extract style and content information of given images. In contrast, a prompt template that only contains positive identifiers, \textit{e.g.}, the prompt used in Dreambooth~\cite{dreambooth}, faces two challenging issues: 1) Over-fitting to source or target image. 2) Failed to decouple style and content information. To better characterize the claims, we fine-tune LDM with three different configurations, positive identifiers, positive-negative identifiers, and our positive-negative identifiers plus the auxiliary image set. As shown in Figure~\ref{fig:aba1}, the results generated by positive identifier overfit to source image and often generates ``haloing'' artifacts around faces. Adding negative identifiers tends to alleviate this problem and generate acceptable images. Nevertheless, it may fail to capture the local stroke characteristics and generate sharp details, and there still exists entanglement between style and content. In comparison, the full prompt template achieves the best performance in terms of style effect and content preservation.

\noindent\textbf{{\textit{Effect of Fine-grained Content Controller.}}}
To verify the effectiveness of the proposed FCC module, we perform the evaluation by inferring artistic portrait without it. As shown in Figure~\ref{fig:aba2}, StyO without FCC will generate diverse artistic portraits that maintains source identity, due to the content identifiers also store the content information of source image. This characteristic allows users to have more choices, such as different poses, different hairstyles and eye colors.  With the introduction of FCC module, it is obvious that the diversity of syntheses is significantly reduced and the identity fidelity to source image is also improved. 

\noindent\textbf{{\textit{Hyper-parameters in Augmented Text Prompt.}}}
The ablation in Figure~\ref{fig:aba3} analyzes the visual results by varying the number of style identifiers $n_{s}$ and the number of content identifiers $n_{c}$ in FCC. Here, we find that with the increase of $n_{s}$, the style of syntheses gradually change from photorealistic to animated. Similarly, a larger $n_{c}$ will force the output image to ``overfit'' to the source image. Actually, there is a trade-off between content faithfulness and style similarity related to hyper-parameter $n_{s}$ and $n_{c}$. Note that in practice, we can experiment with various hyper-parameters to synthesize multiple artistic portraits, allowing the user to choose among different levels of artistic effects.
\vspace{-2mm}

\section{Conclusions}
In this paper, we present a novel StyO model to effectively solve the one-shot face stylization. StyO exploits a disentanglement and recombination strategy. For this purpose, StyO introduces two new modules, \textit{i.e.}, Identifier Disentanglement Learner (IDL) and Fine-grained Content Controller (FCC). IDL aims to disentangle style and content attributes of images into different identifiers and FCC recombines style and content identifiers to form prompts describing the stylization faces. Extensive results show that our StyO model can generate high-quality stylization results in one-shot manner, clearly outperforming existing methods. 

\section*{Acknowledgements}
This paper is supported is supported by National Key R\&D Program of China (2021YFA1000403), National Natural Science Foundation of China (Nos. U23B2012, 11991022), and Fundamental Research Funds for the Central Universities (E3E41904X2).

\bibliography{aaai25}

\end{document}